\documentclass[runningheads]{llncs}
\usepackage{graphicx}
\usepackage{hyperref,xcolor}
\usepackage{biblatex}
\begin{document}
\title{The WebCrow French Crossword Solver\thanks{Supported by expert.ai, \url{https://www.expert.ai/}}}
\author{Giovanni Angelini\inst{1} \and
Marco Ernandes\inst{1} \and
Tommaso Iaquinta\inst{2} \and
Caroline Stehlé\inst{3} \and
Fanny  Simões\inst{4} \and
Kamyar Zeinalipour\inst{2} \and
Andrea Zugarini\inst{1} \and
Marco Gori\inst{2}
}
\authorrunning{G. Angelini et al.}
\institute{expert.ai, Modena, Via Virgilio, 48/H – Scala 5 41123, Italy 
\and
Università degli Studi di Siena, Via Roma, 56, 53100 Siena, Italy
\and
Université Côte d'Azur, INRIA, Institut 3IA Côte d'Azur, France
\and
Université Côte d'Azur, Institut 3IA Côte d'Azur, France}

\maketitle              % typeset the header of the contribution

\begin{abstract}
Crossword puzzles are one of the most popular word games, played in different languages all across the world, where riddle style can vary significantly from one country to another. 
Automated crossword resolution is challenging, and typical solvers rely on large databases of previously solved crosswords. 
In this work, we extend WebCrow 2.0, an automatic crossword solver, to French, making it the first program for crossword solving in the French language. 
To cope with the lack of a large repository of clue-answer crossword data, WebCrow 2.0 exploits multiple modules, called experts, that retrieve candidate answers from heterogeneous resources, such as the web, knowledge graphs, and linguistic rules. 
%It is the first application of its kind for French crosswords and it is designed to possibly solve any crossword puzzle which uses Latin characters.
%WebCrow 2.0 follows a three-phase approach: clue analysis and answering, belief propagation and grid filling. 
%WebCrow 2.0 is able to incorporate various clue-answering experts, also developed outside WebCrow itself such as NLP experts, rule-based, web search, and crossword expert modules. NLP experts leverage Expert.ai NLP technologies, including Knowledge Graphs, and sentence embeddings technologies for finding meaningful answers to clues. The generated candidate answer lists are ordered based on probability and confidence scores are given. A merger system combines the lists.
%Grid filling treats crossword solving as a Probabilistic-CSP problem. It makes use of a belief propagation strategy, using the puzzle contrarians\cite{littman2002probabilistic}, then it fills the grid letter by letter with a Char Based Solver algorithm.
We compared WebCrow's performance against humans in two different challenges. Despite the limited amount of past crosswords, French WebCrow was competitive, actually outperforming humans in terms of speed and accuracy, thus proving its capabilities to generalize to new languages.

\keywords{Natural Language Processing \and Crossword \and Crossword Solver \and Artificial intelligence \and Linguistic Puzzles \and Probabilistic Constraint Satisfaction} 
\end{abstract}
\section{Introduction}
Crossword puzzles have gained immense popularity as a widely played language game on a global scale. Daily, millions of individuals engage in the challenge, requiring a combination of skills. 
%that encompass a sharp comprehension of often ambiguous or tricky clues, a vast knowledge base, and an astute heuristic approach to accurately fill in the puzzle. \\
To solve crosswords effectively, humans need to possess a broad vocabulary, general knowledge across various subjects, and the ability to decipher wordplay and puns. Human solvers should master the crossword language, its peculiarities, and specific knowledge belonging to the country in which it is spoken. They must also excel in pattern recognition, interpret contextual clues accurately, employ problem-solving strategies, and demonstrate patience and perseverance. Mastering these skills enables individuals to tackle crossword puzzles with efficiency, accuracy, and a higher likelihood of success.\\
This scientific paper introduces a novel version of WebCrow 2.0, an AI-powered application specifically designed for efficiently solving French crosswords. It represents the first of its kind in the realm of French crosswords, building upon the previous versions developed for Italian and American crosswords. We will discuss the peculiarities of the French version in section \ref{sec:fr} and the underlying architecture in section \ref{sec:system}.\\
Solving crosswords based on clues is widely recognized as an AI-complete problem\cite{littman2001computer}, owing to its intricate semantics and the extensive breadth of general knowledge required. Artificial intelligence has recently shown an increasing interest in crossword solving. \cite{wallace2022automated} Through this work we are introducing a notable milestone in the literature, the French WebCrow system, which achieved human-like performance on French crosswords by leveraging numerous knowledge-specific expert modules.\\
WebCrow 2.0 can rely on a limited amount of previously solved crosswords and clue-answers pairs. In the case of French crosswords, WebCrow 2.0 made use of about 7.000 previously solved crossword puzzles and about 312,000 unique clue-answers pairs. Studies in American crosswords rely on millions of clue-answers pairs, 6.4M \cite{wallace2022automated}, and on the fact that almost all of the answers are in previously seen crosswords. This is not the case with French crosswords, for which the availability of a huge collection is limited, thus a more robust approach is required.\\
The primary objective of French WebCrow is to establish its competitiveness against human crossword solvers by leveraging expert modules, NLP (Natural Language Processing)  technologies, web search, and merging techniques to efficiently generate candidate answer lists and fill crossword grids accurately. The goal of the web search source of information is to provide accurate solutions to crossword puzzles without the burden of maintaining an up-to-date multitude of domain-specific modules. By tapping into the web as an extensive source of information, French WebCrow offers the promise of scalability and adaptability.\\
The upcoming sections provide information on related works and a comprehensive overview of the various components of WebCrow 2.0. Detailed explanations will be given on the French WebCrow version, accompanied by a thorough analysis of the experimental results. Finally, the paper will conclude by summarizing the findings and highlighting the significance of this research in the field of crossword solving.

\section{Related Works}
In the literature, various attempts have been made to solve crossword puzzles. However, none of these approaches have adequately addressed the specific challenges posed by French crosswords. In the following, we will delve into a review of existing works that have tackled the task of solving crosswords.\\
One of the first works on crossword solving is Proverb\cite{littman1999solving}, which tackles American crosswords. The system makes use of independent programs that solve specific types of clues, leveraging information retrieval, database searching, and machine learning. During the grid filling phase, it tries to maximize the number of most probable words in the grid, using a loopy belief propagation, combined with A* search \cite{Hart1968}.\\
Taking into account the Proverb experience, WebCrow \cite{angelini2005solving, ernandes2005webcrow, angelini2005webcrow} is the first crossword solving for Italian crosswords. WebCrow introduces the use of a Web Search Module (WSM), that extracts and filters potential answers from the Web, being this an extremely rich and self-updating repository of human knowledge. Additionally, the system retrieves clues from databases of previously solved crossword puzzles (CPs). A merging process aims to consolidate the potential solutions from both web documents and previously solved CPs. Subsequently, the system employs a probabilistic Constraint Satisfaction Problem (CSP) approach, similar to the Proverb system \cite{littman2002probabilistic}, to fill the puzzle grid with the most suitable candidate answers. Both Proverb and WebCrow proved to be better-than-average cruciverbalists (crossword solvers).\\
Following these experiences, we can find Dr.Fill work\cite{ginsberg2011dr}, a program designed to solve American-style crossword puzzles. Dr.Fill converts crosswords into weighted Constraint Satisfaction Problems (CSPs) and utilizes innovative techniques, including heuristics for variable and value selection, a variant of limited discrepancy search, and postprocessing and partitioning ideas. The program's performance in the American Crossword Puzzle Tournament suggests it ranks among the top fifty crossword solvers globally.\\
In the field of crossword solving, there is also SACRY\cite{barlacchi2015sacry}, introduced in 2015, a system that leverages syntactic structures for clue reranking and answer extraction. The authors build upon the foundation of WebCrow \cite{angelini2005solving, ernandes2005webcrow, angelini2005webcrow} to develop SACRY. The system utilizes a database of previously solved crossword puzzles (CPs) to generate a list of candidate answers. One of the key contributions of SACRY is its emphasis on exploiting syntactic structures. By incorporating syntactic analysis, SACRY improves the quality of the answer list, enhancing the accuracy of crossword puzzle resolution.\\
Recently, there is the Berkeley Crossword Solver, a cutting-edge approach that revolutionizes automatic American crossword puzzle solving. The system employs neural question-answering models to generate answer candidates for each crossword clue and combines loopy belief propagation with local search techniques to discover complete puzzle solutions. One of the standout features of the Berkeley Crossword Solver is its use of neural question-answering models, which significantly enhances the accuracy of generating answer candidates. \\
In the subsequent sections, we will provide a comprehensive and detailed explanation of the various components comprising our system. We aim to delve into each part, elucidating its functionalities and intricacies, to offer a thorough understanding of our system's architecture and its underlying mechanisms.

\section{Overview of WebCrow 2.0}
WebCrow 2.0 is based on the previous WebCrow project experience(\cite{ernandes2005webcrow}). 
As shown in Fig.1, WebCrow has a first phase of clue analysis and clue answering. For each clue a list of candidate answers, of the
suitable length, is generated by a variable number of experts. Then, all ordered lists are merged into a unique list for each clue. The merging phase takes into account information like the expert module's confidence, the clue type and the answer length. The list merger module and list filtering module, based on morphological information, are both trainable on data.
Next comes a belief propagation step(\cite{littman2002probabilistic}) which reorders the candidate lists based on the puzzle constraints. Finally, the last step is the real-solving mechanism that actually fills the grid with letters, using a new grid-filling approach, the Char Based Solver algorithm.

\begin{figure}[htp]
    \centering
    \includegraphics[width=12cm]{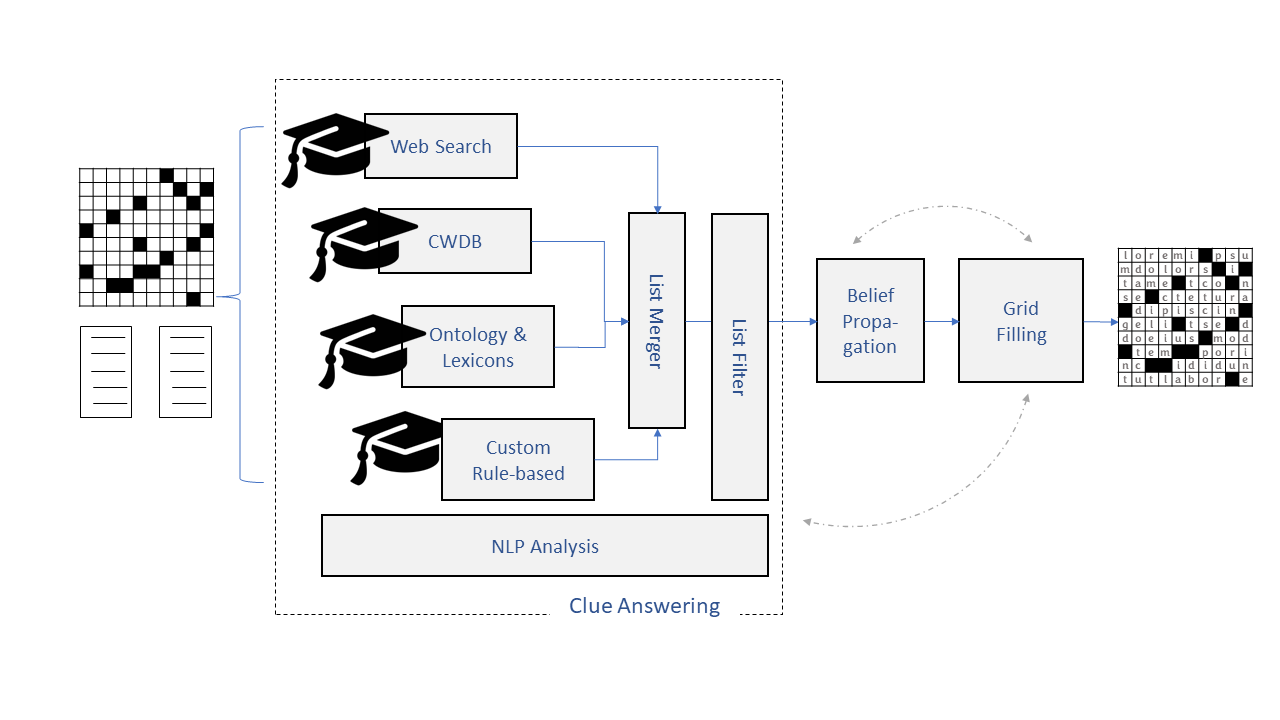}
    \caption{WebCrow Overview.}
    \label{fig:WebCrow_overview}
\end{figure}

\subsection{Modularity}
WebCrow 2.0 has a modular architecture, based on Redis as a communication backbone. Redis implements a Publish/Subscribe messaging paradigm which allows asynchronous communication between agents of nearly every programming language \cite{redis:pubsub}.
The advantage is that with little effort we are able to design expert modules for new languages or based on state-of-the-art natural language processing techniques. 

Based on our experience, expert modules should cover these three types of knowledge:
\begin{itemize}
\item \textbf{Lexical and Ontological Knowledge}: knowledge about the way we use language to represent the world and organize information.
\item \textbf{Crossword-specific experiential Knowledge}: frequent crossword clue-answer pairs, specific conventions
and rules which recur in  crossword puzzles.
\item \textbf{Factual and Common Knowledge}:  encyclopedic knowledge, common sayings, facts, and events of a common cultural background. The Web can be viewed as a repository of this kind of knowledge.
\end{itemize}

In the next section, we are going to analyze in more detail the most crucial expert modules that contribute to the creation of candidate answer lists.

\subsection{The Expert Modules}

\subsubsection{Word Embedding expert}\label{sec:useqq}

The Word Embedding expert takes into account the idea that crossword puzzles often contain knowledge that has already been encountered in previously solved crosswords. Word embeddings \cite{mikolovetal2013distributed,yitanli2015,mikolov2018, devlin2018bert, cer2018universal} offer a way to map individual words or sequences of words (sentences) to specific vectors within a high-dimensional geometric space. This mapping ensures that similar words or sentences are located in close proximity to each other, while sentences with unrelated meanings are positioned far apart.

Building upon a retrieval and ranking approach for crossword clue answers \cite{zugarini2021multi}, this expert employs the Google Universal Sentence Encoder (USE) to embed each puzzle clue. It then searches for the most similar clues within the clue-answers dataset, leveraging the capability of word embeddings to discover linguistic connections between clues.

\subsubsection{WebSearch expert} \label{sec:ws}
The Web Search Module utilizes web documents and search engines to identify suitable answers for crossword clues. It consists of a web-based list generator, a statistical filter, and an NLP category-based filter. The module excels in handling longer word or compound word targets. It is particularly useful for obtaining up-to-date data that may not be available in other modules. In our current implementation, we have seamlessly integrated the Bing API\cite{bingapis}, but it is also feasible to utilize alternative search APIs.

\subsubsection{Knowledge Graph expert}\label{sec:kge}
In this paper, we introduce a novel expert that utilizes expert.ai's linguistic knowledge graph\cite{expertaikg}, which provides a domain-independent representation of the real world through concepts and their related meanings and the different relationships that exist among concepts. Each linguistic concept is explained using its similar meanings, its definition, and its related concepts extracted from the Knowledge Graph. The concept is then mapped using word embedding, which enables a search similar to the Word Embedding expert. \\
By employing word embedding techniques, the concept can be effectively searched, similar to the functionality of the Word Embedding expert. This new expert has proven to be invaluable in solving clues that require both lexical and ontological knowledge, such as ``Sick'' [ILL] or ``Supportive kind of column'' [SPINAL]. Inside expert.ai Knowledge Graph "sick" and "ill" are two words belonging to the same concept, they are synonyms. As far as "spinal", there is a concept "spinal column" which is a specification (kind of) of the concept "column". 
%Our results demonstrate the effectiveness of the Expert.ai Knowledge Graph Specialist in solving crossword puzzles and other language-based tasks. We believe that this expert has the potential to revolutionize natural language processing and contribute to the development of more advanced AI systems.

\subsubsection{Other Expert Systems for Language-Specific Crosswords}
Expert systems for language-specific crosswords are designed to cater to the specific nuances of the language. For example, in Italian crosswords, there are often word plays with 2-letter answers. To address this, a hard-coded expert system has been developed that encodes many of the possible types of word plays, resulting in high-confidence answers. A similar approach has been taken for French solvers, as described in Section \ref{sec:rb}. However, such a situation is not present in American-style crosswords, where the minimum number of letters for an answer is 3.

\subsection{Merging}
Once all the experts have produced their outputs, which are lists of candidate words each one associated with a probability the list is merged together in a unique list. The merging procedure consisted of a weighted average of the experts list based on the length of the answer, the weights are picked based on a specific training phase.

\subsection{Grid Filling}\label{sec:grid_filling}
 
For the grid-filling phase, we made use of a Char Base Solver. This approach is more robust in case some candidate lists do not have the correct answers, which is very likely in French crosswords.\\
For each slot $s$ we cumulate the probability mass $p^s_d(c)$ of a letter $c$, in a given direction $d$ (Across or Down), adding all the probabilities of words that contain letter $c$ in the slot $s$ with direction $d$. We compute the probability mass $p^s(c)$ as:

\begin{equation}
p^s(c) = p^s_{\mathit{A}}(c) \cdot p^s_{\mathit{D}}(c), 
\end{equation}

This can be seen as the probability of the letter $c$ being correctly inserted in a given cell, considering the constraint network and the answer lists. We then use two criteria to assign to the given box the letter $c$ and in this way constrain the grid filling.

\begin{equation}\label{eq:crit1}
    \left(p^s(c) > 99.99\% \right)  and \left(\mathit{best}_\mathit{A}(c) == \mathit{best}_\mathit{D} (c)\right)\\
\end{equation}
\begin{equation}\label{eq:crit2}
\left(p^s(c) > 99.00\% \right) and
\left(\mathit{best}_\mathit{A}(c) == \mathit{best}_\mathit{D} (c)\right) and\\
\left( p^s_{\mathit{A}}(c),p^s_{\mathit{D}}(c) > 90\% \right)     
\end{equation}

In other terms equation \ref{eq:crit1} states that a letter $c$ is chosen for a cell if the confidence on that letter being in that cell is higher than 99.99\% and it is the most likely prediction in both directions. Where $\mathit{best}_\mathit{A}(c)$ is the most likely letter in the across direction and $ \mathit{best}_\mathit{D}(c)$ the most likely in down direction. Obviously this two letter must be the same.

Equation \ref{eq:crit2} instead states that if the confidence on a given letter being in a given cell is only 99.00\% then it is not enough to be the most likely for both directions ($\mathit{best}_\mathit{A}(c) == \mathit{best}_\mathit{D}(c)$) but that letter must have more than 90\% probability for both directions. 

If either of these criteria is met, then the character is assigned to that particular position. Otherwise, it will be filled in a second phase with the most probable word that does not break any other char-based constraint. In the unlikely event that no word satisfies the bond, the cell is left unfilled or could be filled by another post-processing expert, such as an implicit module.

\section{The French Crosswords}\label{sec:fr}

\subsection{Format and Rules}
The French crossword format is similar to Italian crosswords. Unlike American crosswords, two-letter words and ``Blind cells'' (cells that belong to only one word) are allowed. Stacked answers made up of multiple words are less common in French crosswords and generally correspond to expressions. \\
French crossword puzzles vary greatly in size and in the type of knowledge used. In the next sub-sections, we will describe in more detail these aspects.

\subsection{French Crosswords Dataset}\label{sec:fr_dataset}

\begin{figure}
    \begin{center}
        \includegraphics[width=\textwidth]{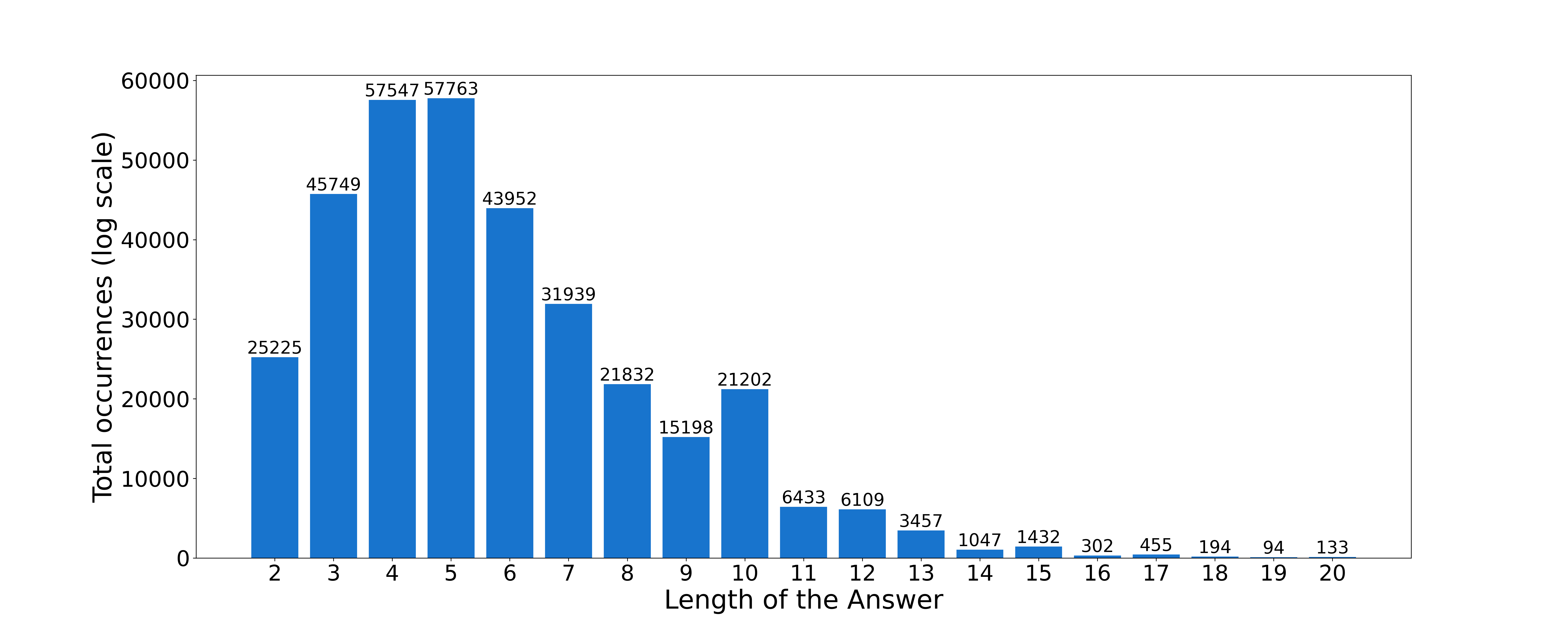}
    \end{center}
\caption{Statistics on French crosswords dataset.} \label{fig:answer_len}
\end{figure}

For the French dataset, we collected over 300,000 clue-answer pairs, with the answer length distribution shown in Figure \ref{fig:answer_len}. Additionally, we compiled a collection of approximately 7,000 solved crossword puzzles from diverse sources. We owe our success in this endeavor, completed in just a few months, to the invaluable collaboration of two prolific authors, Serge Prasil and Michel Labeaume.

\begin{table}
\begin{center}
\caption{Dataset of previously seen clue-answers pairs and crosswords.}\label{tab_dataset}
\begin{tabular}{|l|c|c|}
\hline
{\bfseries Language} &  {\bfseries\ unique clue-answers pairs} & {\bfseries\ crosswords} \\
\hline
{\bfseries American Crosswords}            &  3,100K    & 50,000\\
{\bfseries Italian Crosswords}   &  125K   & 2,000\\
{\bfseries French Crosswords}    &  300K   & 7,000\\
\hline
\end{tabular}
\end{center}
\end{table}

As we can see in table \ref{tab_dataset}, the French dataset of previously seen clue-answers pairs and crosswords is comparable to the Italian dataset, while the American dataset is considerably huger. Moreover,  American crosswords are more standard. Almost all clue answers are present in previous crosswords, which is not the case with French crosswords.\\
In figure \ref{fig:answer_len} we show the statistics of the answer length present in French crosswords. The majority of the answer's lengths are below 10. Answers with higher lengths are covered by verb inflections, compound words, or linguistic expressions.

\subsection{Linguistic and Cultural Peculiarities}
Unlike Italian and American crosswords, French crosswords use a wide range of verb inflections in their solutions, covering nearly every possible tense and person. However, the definitions provided in the clues often lead to the correct inflection.\\

Furthermore, we have observed that French crossword authors have distinct individual styles that vary greatly from one another. As in other crossword languages, the aim of a crossword author is to provide clues that are obscure enough while keeping solutions that should appear obvious once found \cite{berthelier2018humour}. He must find the right level of difficulty for all the pairs of solutions. When this level is too high, the risk is to discourage people from trying to solve the crossword. On the contrary, if the clues are too simple, it is a memory or patience game, but there is no challenge, and usually, French crossword players prefer tricky enigmas, with few clues, twisted words, or traps. \\
French crossword authors inherit from the art of conversation in classical French culture, which is well represented by the periphrase ``la langue de Molière'' to designate French. As a result, French authors take pride in being witty in the definitions they provide. They must be creative in finding jokes that make the solver laugh \cite{berthelier2018humour}, which leads to the development of distinct individual styles.

\subsection{Examples of clues in French crosswords} \label{sec:ex_clues}

In this section, we categorize the types of clues found in French crosswords and provide illustrative examples. Some of the examples are very specific to the French language, in particular the examples given in sections Inflections or Domain Specific Knowledge, and some other examples related for instance to rare words or word games can be found in other languages as well.

\subsubsection{Inflections}
French crosswords make extensive use of rare verb tenses and modes, which can make it challenging to find the correct inflection of the word to be guessed through a direct web search. For instance, in the following clue answer pair: \textit{Auraient des soucis excessifs [CAUCHEMARDERAIENT]}, the verb to guess ``cauchemarder'', which means ``having a nightmare'', is rarely used at the conditional present, at the third person plural.
In another example, \textit{Apitoie [ATTENDRISSE]}, the clue can refer to either the first or third person at the indicative or subjunctive present tense. Depending on the verbal group of the solution, the inflection can vary significantly at these tenses and persons.

\subsubsection{Rare words}
Some clues may involve words that are rare in French, either because they are ancient words or foreign words, or these words belong to the literary register or, conversely, to the colloquial or slang register. For instance, the solution of the clue \textit{Dessiner sans soin [STRAPASSER]} is an old verb. As the frequency of these words is low, they may appear with a very low probability, and in some cases, they may not appear at all in the candidate solutions list.

\subsubsection{Domain Specific Knowledge}\label{domain_spec_kg}
Some puzzles require domain-specific knowledge, such as very specific geographical knowledge. For example, a clue may be: \textit{Elle habite une commune située dans le département de l'Isère [SICCIOLANDE]}, meaning that we need to search for the name of the female inhabitants of a city in a specific French department. There is no generic rule in French for determining the name of the inhabitants from the name of the city, and sometimes the name of the inhabitants (in this case, ``SICCIOLANDE'') can be very different from the city name; in this example, the city name is ``Siccieu-Saint-Julien-et-Carisieu''. Therefore, solving this type of riddle requires a combination of encyclopedic knowledge, spelling rules, and potential knowledge of spelling exceptions.

The following example requires specific knowledge of French literature: \textit{Le bleu et le blanc du poète [OE]}. This example pertains to the poem ``Voyelles'' by the renowned French poet Arthur Rimbaud, where each vowel is linked to a color. In this poem, the vowel ``O'' is associated with the color blue (``bleu''), and the vowel ``E'' is associated with the color white (``blanc'')

\subsubsection{Generic Words With Few Indices} 

On the other hand, some clues may consist of a few generic words such as color names and adverbs, which can be linked to numerous solutions. In such cases, the definition is not clearly connected to the answers, making automatic graph search more challenging.
For instance, consider the following clue: \textit{Pétales de rose [ESE]}. One may be misled by the words ``Pétales'' and ``rose'', which could refer to the lexical field of flowers. However, in French crosswords, they refer to the compass rose, and the solution could be of the type ESE (``Est, Sud, Est'' meaning direction East, South, East), NN, NSN, and so on.

\subsubsection{Word Games}
  
Word games are a type of clue in which the solver must manipulate the multiple meanings of the words in order to arrive at the solution. In crossword puzzles, common word games involve the letters of a single word, which may be either part of the clue or part of another word that must be guessed. For example, consider the clue \textit{A la sortie de Strasbourg [RG]}. The phrase ``A la sortie de'' translates to ``At the exit of'' and suggests that the solution is composed of the last letters of the word ``Strasbourg''. This clue is made more challenging by the fact that ``Strasbourg'' is a proper noun, and solvers may be tempted to look for a solution that is geographically related to the city.

\subsubsection{Two Steps Clues}
Some crossword puzzles can be challenging as they require two or more steps to arrive at the solution. For instance, consider the clue \textit{À l'envers : coût [FIRAT]}. To solve this puzzle, one must first identify a synonym for the word ``coût'' (TARIF) and then invert the letters (FIRAT), as indicated by the phrase ``À l'envers :''. Similarly, in the clue \textit{Grecque a l'envers [ATE]}, the solver must recognize that ``Grecque'' refers to a Greek letter before inverting the letters of the word found. In the example \textit{Impro de jazz sans voyelle [SCT]}, while it may seem straightforward to humans, this could prove to be a challenging task for a machine. The solver should find the answer to the definition of ``impro de jazz'' (``jazz improvisation'') without any information about the word length before removing the vowels.

\subsubsection{Multiple Categories}
Finally, crossword puzzles often combine multiple difficulties. In this example:  \textit{Attaquerai les portugaises [ESSORILLERAI]}, the author Serge Prasil used slang expression ``les portugaises'', to refer to ears. The verb to guess is further an ancient word, a medieval torture that means cutting off the ears, in an unusual form, because it is conjugated at the future.

\section{The System Architecture} \label{sec:system}

 The recent changes in the architecture allowed for easy incorporation of new agents and modification of existing ones by simply adjusting the parameter configuration. For example, the web-search expert (see Section \ref{sec:ws}) was ported to French by modifying the query language in the parameter set.

To update the Word Embedding Expert, we required the French crosswords dataset described in Section \ref{sec:fr_dataset}. The clues had to be encoded further with the Universal Sentence Encoder, as explained in the Word Embedding expert section (see Section \ref{sec:useqq}).

After implementing these two expert agents, we analyzed the results to identify the areas where most errors occurred. We discovered that 29\% of missing answers were due to missing verb inflections, and 8\% were due to adjective or noun inflections. Among all verb forms, the present tense was used only 20\% of the time, while the past simple, a tense rarely used in everyday life, was used 40\% of the time. Among the inflections of adjectives, the feminine form was used 58\% of the time, and the plural form was used 55\% of the time.

\subsection{Knowledge Graph Expert}

As per the analysis of the most common errors, we have enhanced expert.ai's French knowledge graph. The results analysis revealed the need to incorporate inflections of verbs, adjectives, and nouns. To achieve this, we followed the same approach as described in Section \ref{sec:kge}. However, in this case, in addition to adding the connected concepts with the same description, we also included the required inflections.

\subsection{Lexicon}

In addition, we identified a need to enhance the lexicon utilized by WebCrow. To address this, we incorporated Lexique 3.83, a French lexicon database containing approximately 123K distinct entries of at least 2 letters, as described in \cite{pallier2019openlexicon}. We combined this dataset with data from a French dictionary, resulting in a final lexicon comprising approximately 198K words.

\subsection{Rule-Based Expert}\label{sec:rb}

We have developed a Python-based expert module for French crosswords that can decipher common word games. The module is designed to identify target words in the clues and provide associated lists of solutions. The target words may include Arabic number conversions to words, Roman numerals, chemical elements from Mendeleev's table, French departments, grammar lists (such as personal pronouns, conjunctions, and prepositions), and Greek letters.

Furthermore, the Rule-based expert was designed to decipher clues that indicate the presence of word games in finding the answers, and where the solution involves either the inversion of a word, a reduced set of letters, or a mix of letters. The word on which the word game applies may be included in the clue or not. In the latter case, which we called ``two steps clues" in chapter \ref{sec:ex_clues}, the rule-based expert first searches for a list of possible solutions by calling the Word Embedding expert and then applies the word game to the letters of each word in the list.
  
\section{Experimental Results}
In this section, we present the comprehensive results obtained from our experimentation. Following the development of the system, as outlined in the preceding sections, we proceeded to assess its performance on previously unseen crosswords. 

\subsection*{Test Dataset}
To ensure a robust evaluation, we carefully selected a dataset comprising 62 distinct crosswords that were published subsequent to the crosswords used for constructing the different experts, such as the Word Embedding expert\ref{sec:useqq}. This selection criterion ensured that there was no overlap between the crosswords utilized for training and those employed for testing purposes.\\
To evaluate the performance of our proposed solution, we conducted an extensive analysis using a diverse set of crossword puzzles sourced from multiple authors and publications. Our dataset comprises 10 puzzles each from two renowned creators, Michel Labeaume and Serge Prasil. Furthermore, we incorporated 40 additional crosswords from established publishers to facilitate a thorough assessment. Detailed information about the test crossword can be found in Table \ref{tab:testset}.\\

\begin{table}[]
    \centering
    \caption{Test CrossWords.}\label{tab:testset}
    \begin{tabular}{|c|c|c|}
    \hline
    \textbf{Source} & \textbf{Number of Puzzles} & \textbf{Dimension} \\ \hline
    Michel Labeaume &10 &10 x 10 \\
    Serge Prasil & 10 & 20 x 20\\
    Other Sources & 42 & Variable max 15 x 15 \\ \hline

    \end{tabular}
\end{table}

We used diverse crosswords to test the system's ability to handle different puzzle styles, author preferences, and construction variations. This approach helped us understand the system's performance and adaptability in unseen crosswords.

\subsection*{Results}

We evaluated the system's performance using three distinct metrics: percentage of correct words, which measures the accuracy of inserting the correct target answers, percentage of correct letters, which evaluates the accuracy of inserting individual letters, and percentage of inserted letters, which assesses the system's ability to fill crossword slots.\\
For a comprehensive overview of these metrics across different sources of crosswords, refer to Table \ref{tab:testmetrics}. It encapsulates the corresponding results obtained from the test sets of various crossword sources, shedding light on the overall performance of our system in solving French Crosswords.

\begin{table}[]
    \centering
    \caption{Performance of the System on the Test CrossWords.}\label{tab:testmetrics}
    \begin{tabular}{|c|c|c|c|}
    \hline
    \textbf{Source} & \textbf{Words Accuracy} & \textbf{Letters Accuracy} &  \textbf{Inserted Letters}\\ \hline
    Michel Labeaume & 92.97\% &98\%& 100\% \\
    Serge Prasil & 91.82\% & 96.9\%& 99.15\%\\
    Other Sources & 73.86\% & 81.16\%& 96.99\% \\ \hline

    \end{tabular}
\end{table}

 Our crossword solver achieved impressive results in solving French crosswords from Michel Labeaume and Serge Prasil, with some 100\% solved crosswords. On the other sources, the performance varied a lot, we had some sources with fully correct solved crosswords, while on other crosswords the system performed poorly. Based on our analysis some authors use very specific styles and knowledge, which demonstrates that solving crosswords is an AI-complete and open-domain problem.
 In some cases, answers were very domain-specific, see section \ref{domain_spec_kg}.\\

Overall, these remarkable results demonstrate the robust performance of our system in solving French crosswords. The accuracy rates obtained highlight the system's ability to effectively fill in words and letters, thus confirming its competence in solving French crossword puzzles.\\

In table \ref{tab_ablation_test}, we tested the system by removing some expert modules. These tests show that each module is necessary to obtain the best results, the Full version, and that different source of knowledge is required to solve crosswords. Unlike American crossword studies, there is not a huge dataset of previously solved crosswords. Moreover, French crosswords are not as standard as American ones. Each crossword can vary a lot, influenced by the style and imprint of its author.

\begin{table}
\begin{center}
\caption{Ablation test.}\label{tab_ablation_test}
\begin{tabular}{|l|c|c|c|}
\hline
{\bfseries Configuration} &  {\bfseries\% of correct words} & {\bfseries\% of correct letters} & {\bfseries\% word drop}  \\
\hline
{\bfseries Full}            &  65.71    & 75.22     & -\\
{\bfseries No Rule based}   &  65.16    & 74.79     & 0.55\\
{\bfseries No Websearch}    &  61.60    & 72.68     & 4.11\\
{\bfseries No Lexicon}      &  61.36    & 71.98     & 4.35\\
{\bfseries No KG}           &  56.28    & 68.38     & 9.43\\
\hline
\end{tabular}
\end{center}
\end{table}

To gain insights into our system's strengths, limitations, and relative performance compared to human crossword solvers, we conducted challenging competitions. The subsequent section presents a detailed analysis of these comparative evaluations.

\subsection{AI vs Human challenges}

We organized an internal challenge at INRIA to evaluate our system's performance in a real-world scenario, putting it against human participants. The challenge included French and American crossword puzzles. Both humans and WebCrow were allowed to utilize web searches during the challenge. \\
The challenge included three crosswords: an easy-medium-level French crossword with a 10-minute time limit (score counted), a medium-hard level French crossword with a 20-minute time limit (score counted), and an American crossword with a 10-minute time limit (score counted). The experimental results, including the performance of WebCrow (Live and Lab), the average human performance, and the best human performance are presented in Table \ref{tab:resultsinria}.

\begin{table}[]
    \centering
    \caption{Results of the Crossword Solving Competition (INRIA).}\label{tab:resultsinria}
    \begin{tabular}{|c|c|c|}
    \hline
    \textbf{Player} & \textbf{ Score } & \textbf{ Time (sec.)} \\ \hline
    WebCrow Live&   296.18 &419\\
    WebCrow Lab & 313.75 &556\\
    AVG Human  & 50.39 & 2570\\
    Best Human & 104.22 & 2700\\
    \hline

    \end{tabular}
\end{table}

Two modes were implemented: ``WebCrow Live'' where the system ran in real-time with predetermined configurations, and ``WebCrow Lab'' where results were computed in advance in the laboratory. It is important to note that variations in web information could lead to discrepancies between the results of the two modes.\\

We also conducted a public challenge at the World AI Cannes Festival 2023, evaluating the French version of WebCrow. There were three challenges, one for each language: French crosswords, Italian crosswords, and American crosswords. Each challenge had two crosswords valid for the competition with time limits. The two French crosswords were created specifically for the challenge by renowned authors Serge Prasil and Michel Labeaume.\\ 

The scoring system gave points from 0 to 100 based on the percentage of correct words (0 to 110 for the second crosswords. Then some additional points (maximum 15) were added based on the percentage of time not used. We had 15 minutes for the first crossword and 20 minutes for the second. Finally, in case of a fully correct answer, 15 points were awarded. \\

\begin{table}[]
    \centering
    \caption{Results of the French Crossword Solving Competition (WAICF).}\label{tab:resultswaicf}
    \begin{tabular}{|c|c|c|}
    \hline
    \textbf{Player} & \textbf{Score} & \textbf{Time (sec.)} \\ \hline
    WebCrow Live&   228,90 &559	 \\
    WebCrow Lab & 249,86	 &368\\
    AVG Humans & 24.24 & 2570\\
    Best Human & 69,53 & 1493\\
    \hline
    \end{tabular}
\end{table}

The detailed experimental outcomes of the WAICF French crossword-solving challenge can be found in Table \ref{tab:resultswaicf}.
This challenge provided insights into WebCrow's performance and its cross-lingual capabilities. Humans cruciverbalist are strong only on one language.\\ 
In the French crossword challenge, there was no strong human competitor present. This leaves space for further challenges with French experts in crosswords.
%It clearly demonstrates that the French WebCrow outperformed human participants in both crossword score and solving time. These findings highlight the exceptional proficiency and efficiency of the French WebCrow, surpassing humans in accuracy and speed. This success has significant implications for the advancement of automated crossword-solving systems, particularly in handling linguistic challenges and cultural nuances. The results provide compelling evidence for the potential of artificial intelligence and natural language processing techniques to effectively tackle complex language-based problem-solving tasks, outperforming humans in both solution quality and speed.

\section{Conclusion and Future Works}
In conclusion, this work represents a significant advancement in the field of crossword solving. By capitalizing on our previous experience in the field we present a novel version of WebCrow 2.0 and its French WebCrow version, which represents the first French crossword solver. \\
In this work we collected a dataset of French crosswords, enabling us to make some comparisons with crosswords in other languages, Italian and American. Moreover, we analyzed the peculiarities of French crosswords. French crossword puzzles vary greatly, they are not standard like the American ones, the size, the knowledge, and the language games involved are influenced by the style and imprint of its author. \\
French WebCrow is an above-human-average crossword solver, but there is still room for improvements. The potential for French WebCrow to achieve competitive performance serves as a strong motivation for further research and development, paving the way for AI-powered crossword solving to reach new heights.\\
There are three main branches for future development. First of all, there is room to improve the performances of both the Italian and French solvers by working on filters and re-ranking based on systems that can predict the grammatical type of the answer. Another improvement can be achieved by leveraging on the output of the Char Based Solver which fills the grid with the most probable letters, leaving empty the cells which have more uncertainty. We would like to implement a system that exploits the letters that are actually fixed to find out the missing ones on the internet or with a Generative Pre-trained Transformer.\\
Another branch of development resides in the intrinsic characteristic of WebCrow 2.0, in which the modularity of its frameworks allows us to add a new language solver with little effort. Of course, as happened for Italian, English, and French, language-specific experts have to be developed to obtain high performances in crossword solving. We are already in touch with German universities to explore this road.\\
The last branch regards the inverse task, the crossword generation \cite{rigutini2012automatic}. The experience gained, but even more the data collected during the WebCrow 2.0 experience, could represent a launch pad for the complex task of crossword generation. Consider that, for instance, the New York Times crosswords (one of the biggest collections of crosswords) contains an average of 96\% of already seen answers, and only the 4\% of the answers, on average, are new \cite{wallace2022automated}. This task is still performed principally through semi-automatic proprietary software. New approaches should take into account Generative Pre-trained Transformers, which, at the moment, represent the most advanced approach for generating text and could be tested on generating crossword clues, which may also be ambiguous or tricky, covering different kinds of human knowledge.\\

\section*{Acknowledgements}
This research owes its accomplishment to the generous collaboration of esteemed French crossword authors, Serge Prasil and Michel Labeaume. The University of Siena, expert.ai, and the 3IA Côte d’Azur Investment in the Future projects administered by the National Research Agency (ANR), under the reference number ANR-19-P3IA-0002, provided invaluable support for this endeavor
\printbibliography
\end{document}